\pdfoutput=1
\documentclass{article}

\usepackage[preprint]{neurips_2024}

\usepackage{fancyhdr}
\usepackage{nth}
\usepackage[utf8]{inputenc} 
\usepackage[T1]{fontenc}    
\usepackage{hyperref}       
\usepackage{url}            
\usepackage{booktabs}       
\usepackage{amsfonts}       
\usepackage{nicefrac}       
\usepackage{amssymb}
\usepackage{microtype}      
\usepackage{xcolor}         
\usepackage{graphicx}
\usepackage{hyperref}
\usepackage{amsmath}
\usepackage{cleveref}
\usepackage{color, colortbl}
\usepackage{tabularray}
\usepackage{multirow}
\usepackage{nicematrix}
\usepackage{caption}
\usepackage{subcaption}
\usepackage{booktabs} 
\usepackage{wrapfig}

\definecolor{mygray}{gray}{0.9}

\title{COOOL: Challenge Of Out-Of-Label\\ A Novel Benchmark for Autonomous Driving}

\definecolor{mydarkblue}{rgb}{0,0.08,1}
\definecolor{mydarkgreen}{rgb}{0.02,0.6,0.02}
\definecolor{darkred}{rgb}{0.8,0.02,0.02}
\definecolor{darkorange}{rgb}{0.40,0.2,0.02}
\definecolor{darkpurple}{RGB}{111,0,255}
\definecolor{myred}{rgb}{1.0,0.0,0.0}
\definecolor{mygold}{rgb}{0.75,0.6,0.12}
\definecolor{mydarkgray}{rgb}{0.66, 0.66, 0.66}

\author{%
   Ali K. AlShami$^{1,\dagger}$, Ananya Kalita$^{2}$, Ryan Rabinowitz$^{1}$, Khang Lam$^{3}$, Rishabh Bezbarua$^{4}$,\\ 
   \textbf{Terrance Boult|}$^{1}$, \textbf{Jugal Kalita}$^{1}$ \\
   $^{1}$Computer Science Department, University of Colorado, Colorado Springs \\
   $^{2}$University of Washington\\
   $^{3}$Information Technology Department, Can Tho University\\
   $^{4}$University of Illinois Urbana-Champaign \\
   \texttt{aalshami@uccs.edu}, \texttt{ananyaka@uw.edu},
   \texttt{rrabinow@uccs.edu}, 
   \texttt{lnkhang@ctu.edu.vn}, \\
   \texttt{rb33@illinois.edu}, 
   \texttt{tboult@uccs.edu}, \texttt{jkalita@uccs.edu}   \\
   \url{https://github.com/alshami52/COOOL_benchmark}
}

\begin{document}

\maketitle

\begin{abstract}
As the Computer Vision community rapidly develops and advances algorithms for autonomous driving systems, the goal of safer and more efficient autonomous transportation is becoming increasingly achievable. However, it is 2024, and we still do not have fully self-driving cars. One of the remaining core challenge lies in addressing the novelty problem, where self-driving systems still struggle to handle previously unseen situations on the open road. With our Challenge of Out Of Label (COOOL) benchmark, we introduce a novel dataset for hazard detection, offering versatile evaluation metrics applicable across various tasks, including novelty-adjacent domains such as Anomaly Detection, Open-Set Recognition, Open Vocabulary, and Domain Adaptation. COOOL comprises over 200 collections of dashcam-oriented videos, annotated by human labelers to identify objects of interest and potential driving hazards. It includes a diverse range of hazards and nuisance objects. Due to the dataset’s size and data complexity, COOOL serves exclusively as an evaluation benchmark. 
\end{abstract}

\section{Introduction}
\label{sec:introduction}

The rapid expansion of autonomous vehicle technology, fueled by AI, machine learning, and sensor systems breakthroughs, holds great promise for improving safety, reducing traffic accidents, and enhancing overall mobility. However, avoiding previously unseen hazards remains a critical challenge for creating truly robust systems. To ensure safer roads, autonomous vehicles must go beyond recognizing common hazards and identify and effectively respond to novel, previously unseen dangers. This necessity has driven the exploration of emerging fields, such as open-vocabulary recognition, to enhance scene understanding and provide more comprehensive explanations.

In recent years, researchers have increasingly explored LiDAR-based approaches to predict future 3D scenes and enhance hazard avoidance \cite{agro2024uno,khurana2023point,weng2022s2net} using datasets like KITTI \cite{geiger2012we} and nuScenes \cite{caesar2020nuscenes}. These methods emphasize point cloud perception and forecasting using self-supervised and unsupervised techniques to capture dynamic 3D environments, enabling real-time scene understanding for autonomous systems. Although LiDAR-based methods show promising results, hazard avoidance is inherently more complex and often requires integrating vision information for more accurate predictions \cite{peri2023towards}. Novel scenarios can become increasingly intricate, making vision information essential. For example, recognizing the color of a traffic light—whether red, yellow, or green—or interpreting road signs cannot be reliably achieved using LiDAR alone. Similarly, understanding pedestrian behavior requires vision-based cues, such as whether a pedestrian is looking at the car or distracted by a device like a phone. Recent research has also begun exploring pedestrian crossing intentions from a psychological perspective \cite{o2022predicting}, emphasizing the necessity of vision-based data for comprehensive hazard detection and prediction.

The COOOL benchmark encourages researchers to explore and address the challenges of \textbf{Novelty} on the roadway. The possible tasks for COOOL benchmark may include detecting Out-of-Label hazards, hazard avoidance in video streams, recognizing hazards in a sequence, predicting hazards from a sequence, using multi-modal models to recognize and predict hazards in the video, and handling low-resolution objects under 50 pixels. These tasks focus on detecting both known and unknown objects in videos, determining which objects in video streams may become hazardous, recognizing hazards and driver reactions in short video sequences, and forecasting potential hazards from image sequences. The benchmark supports approaches for avoiding hazards on the road, particularly solutions inspired by novelty-adjacent areas such as Anomaly Detection, Open-Set Recognition, Open Vocabulary, and Domain Adaptation.

 \section{Related Work}
\label{sec:related_work}
The development of autonomous driving systems has been significantly propelled by creating comprehensive datasets and benchmarks that facilitate the detection and management of road hazards. Notable among these is the KITTI dataset \cite{geiger2012we}, which offers a variety of data modalities, including stereo images and 3D point clouds, to support tasks such as object detection and scene understanding. Similarly, the nuScenes dataset \cite{caesar2020nuscenes} provides a multimodal suite of sensor data, encompassing LiDAR, radar, and cameras, to enhance 3D object detection and tracking capabilities.

Many datasets have been proposed for autonomous driving, with multiple providing onboard forward-facing video footage \cite{liu2024survey}. However, novelty in autonomous driving datasets is still under-explored. While many datasets that address detection in autonomous driving feature a reasonable number of detectable classes \cite{liu2024survey}, evaluating performance with respect to novelty requires diverse and rare sample data which these systems may not have seen labels for in training data.

In novelty detection within autonomous driving, several methodologies have been proposed to identify and respond to unexpected hazards. Amini et al. proposed a novel approach leveraging autoencoders for end-to-end control of autonomous vehicles, integrating mechanisms for novelty detection to address unexpected scenarios \cite{amini2018variational}. The dataset for this study was collected using a Toyota Prius 2015, as part of the MIT-Toyota partnership. Vision data was captured with a forward-facing Leopard Imaging LIAR0231-GMSL camera, featuring a 120-degree field of view and recording 1080p RGB images at approximately 30Hz. 

Greer and Trivedi investigated the use of language embeddings in active learning, with a focus on identifying novel scenes to enhance safety and decision-making in autonomous systems \cite{greer2024towards}. Their study utilizes the LISA Amazon-MLSL Vehicle Attributes (LAVA) Dataset \cite{Kulkarni2021}, which is comprised of traffic signs annotated and bounded in images captured by a forward-facing camera. The dataset includes 10-second video clips providing full scene and trajectory context, along with INS data. Collected from the greater San Diego area, the dataset is carefully curated to represent a diverse range of road types, traffic conditions, weather, and lighting scenarios.

Despite these advancements, there remains a critical need for datasets that specifically address the detection of out-of-label hazards—those not encompassed within predefined categories. The COOOL benchmark aims to fill this gap by providing a diverse collection of dashcam videos annotated to highlight both common and uncommon hazards, thereby offering a valuable resource for evaluating and enhancing the robustness of autonomous driving algorithms in real-world conditions.
\section{COOOL benchmark}
\label{sec:benchmark_evaluation}
\begin{figure*}
    \centering
    \includegraphics[width=12cm]{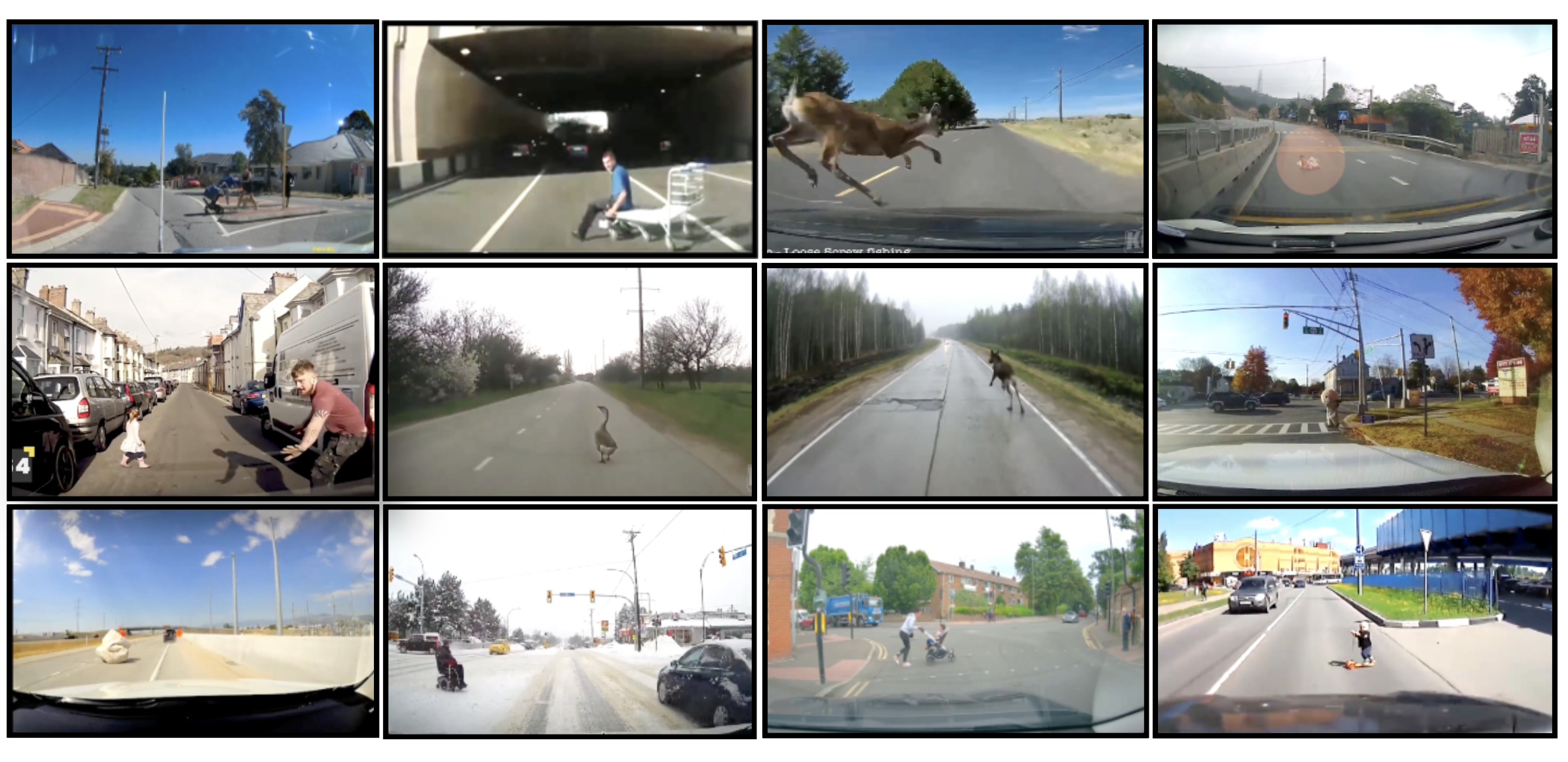}
    \caption{COOOL benchmark examples.}
    \label{figure:COOOL}
\end{figure*}
The COOOL benchmark, 'Challenge Of Out-Of-Label' in Autonomous Driving, is a collection of ~200 high-resolution dashcam videos annotated by human labelers to identify objects of interest and potential roadway hazards. It includes a wide variety of hazards, such as exotic animals (e.g., kangaroos, wild boars), inanimate or unpredictable objects (e.g., plastic bags, smoke), and standard hazards (e.g., cars, pedestrians). COOOL is an evaluation-only benchmark designed to highlight the often-overlooked challenge of detecting out-of-label hazards using vision information in autonomous driving research. Each object is meticulously labeled in every frame, and a unique 'Tag' system provides insights into the vehicle’s movements and the driver’s decisions. The dataset was annotated by undergraduate students from the University of Colorado Colorado Springs (UCCS) and local high school students under the supervision of expert graduate students and professors, using a professional computer vision annotation platform. Examples from the dataset are shown in Figure~\ref{figure:COOOL}.

\section{Workshop Challenge, Baseline, and Evaluation Metric}
\label{sec:data_collection}

\subsection{Challenge}
This competition challenges researchers to address the limitations of current approaches in recognizing and responding to unknown (out-of-label) hazards on the road by proposing new algorithms and systems that advance the field. As autonomous driving becomes a reality and commonplace, we foresee in this dataset the potential for high \textbf{social impact} of being able to respond to novel hazards. Autonomous vehicles that react inappropriately to novel hazards could create life-threatening situations. Researchers from different novelty-adjacent areas can offer diverse approaches to address various aspects of the problem. The current challenge poses three objectives:
\begin{itemize}
    \item Determining when the driver has started to react to a hazard.
    \item Which object(s) in the scene are hazardous.
    \item What is the name of the hazard(s).
\end{itemize}
In the future, the COOOL benchmark can be applied to other tasks, including detecting out-of-label hazards sequences, low resolution hazards, predicting when out-of-label hazards will occur from a sequence of images, and avoidance manuvers for hazards in video streams. These tasks organically build on each other, from detection to prediction to avoidance, approaching a hypothetical vision and possibly vision-language-based autonomous driving application that could give guidance to vehicle planning systems. We provide a baseline solution code addressing each problem in the current challenge, emphasizing the compatibility of the baseline's design in hopes that researchers will extend diverse detection solutions to the more complex issues of prediction and avoidance in future work.

\subsection{Baseline}
We provide a baseline, \href{https://github.com/alshami52/COOOL_benchmark}{Github Link}, that satisfies the three tasks using straightforward methods: reaction to hazards by judging when the driver's state has changed, identifying which object is hazardous, and naming the hazard. In future challenges, we will extend the baseline to encompass more complex tasks.

In the baseline, driver state change is determined using logistic regression to detect when bounding boxes start moving slowly. We predict the hazardous object by determining the closest bounding box to the center of the image. We use a clip interrogator to generate a caption for that object. Advanced models can significantly improve captioning, speed or car orientation detection, and even hazard trajectory prediction.

\subsection{Evaluation Metric}
COOOL competition evaluation metrics are intended to balance the three aforementioned aspects of hazard detection.
We provide systems with a list of bounding boxes and the raw video, which enables diverse approaches to these challenges.
It should be noted that these challenges, while difficult, are only the initial basis for out-of-label hazard detection.
They provide a simple introduction to this vast problem-space, future challenge iterations will involve more complex prediction scenarios.

For predicting which potential hazard is actually hazardous, we simply compute accuracy of predictions over the maximum of ground truth hazards present or number of hazards predicted.
By increasing the total number of hazards by the total number of guesses, we penalize algorithms that over-predict hazards, hoping to get a lucky guess.
We employ the same methodology for hazard descriptions, except we only check whether the class label is in the description, a binary measure.
For driver reaction, we compute accuracy based on ground truth labels for each frame that determine if the driver has reacted to the hazard yet.
Our overall evaluation metric is the macro accuracy of these three measures.
\section{Dataset Analysis}
\label{sec:dataset_analysis}


The COOOL Benchmark is a collection of manually curated and annotated dash-cam videos, highlighting rare or novel hazards on the road as well as driver reactions to these hazards.

\begin{figure}[hbt!]
    \centering
    \includegraphics[width=12cm]{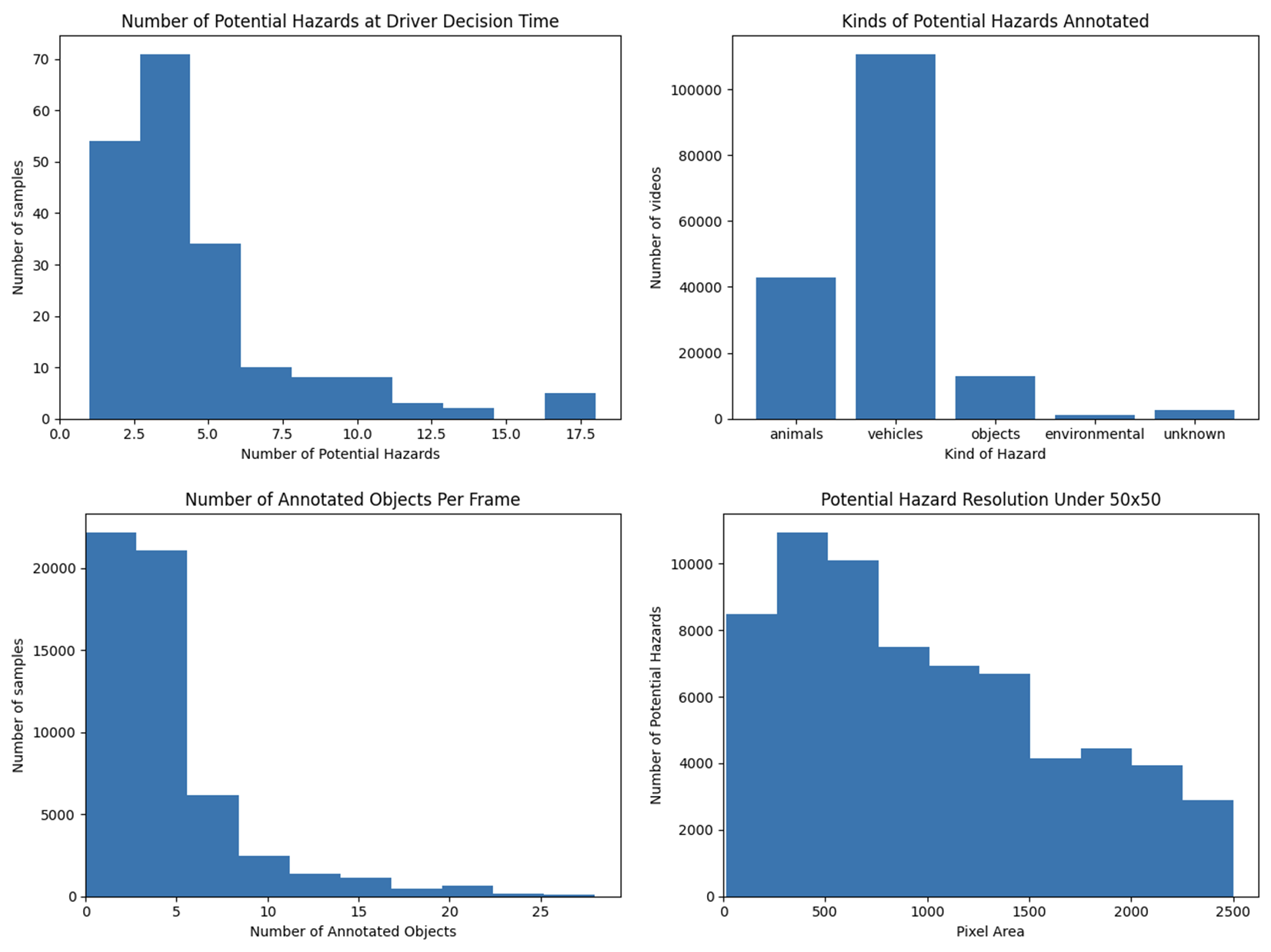}
    \caption{Analysis the COOOL Benchmark}{Four visualizations of COOOL benchmark characteristics describing the distribution of hazard types, low-resolution hazards, hazards at driver reaction moments, and potential hazards per frame.}
    \label{figure:COOOL}
\end{figure}
All videos are captured in High-Definition, providing rich visual information and enabling annotation of distant hazards.
Further, every annotation has been proposed and reviewed by skilled human annotators using state-of-the-art annotation software.
With rich annotations and a diverse collection of real-world dash-cam videos, the COOOL benchmark challenges vision systems to detect out of label hazards with unprecedented detail.

\subsection{Basic statistics}
Analyzing the annotations in the COOOL Benchmark, there are over 100,000 vehicle annotations, ~40,000 animals, as well as a variety of environmental, miscellaneous, and unknown annotations.
The majority of frames contain between 1 and 5 potential hazards, the maximum number of potential hazards per frame is 26.
Over all videos, at the moment a driver reacts to a hazard, there are between 1 and 18 potential hazards, with the majority of driver reactions having 3 potential hazards present.
Out of all annotations, over 10,000 are low-resolution hazards, smaller than 50x50 pixels, this creates particular challenges for early-hazard detection, as distant hazards are small and frequent.
We further analyze these statistics in Figure \ref{figure:COOOL}.

\section{Conclusion}
\label{sec:conclusion}
The COOOL benchmark represents a significant step forward in addressing the challenges of out-of-label hazard detection in autonomous driving systems. By introducing a diverse and meticulously annotated dataset, COOOL offers researchers a valuable tool to evaluate and advance methodologies across tasks like anomaly detection, open-set recognition, open vocabulary, and domain adaptation. Its focus on real-world hazards, including low-resolution and novel scenarios, highlights the critical gaps in current autonomous driving systems and provides a robust foundation for developing safer and more adaptive algorithms. As the field evolves, COOOL is poised to inspire innovative solutions that bring us closer to the realization of fully autonomous and reliable transportation.

\small
\nocite{*}
\bibliographystyle{unsrt}
\bibliography{ref}

\begin{thebibliography}{10}

\bibitem{agro2024uno}
Ben Agro, Quinlan Sykora, Sergio Casas, Thomas Gilles, and Raquel Urtasun.
\newblock Uno: Unsupervised occupancy fields for perception and forecasting.
\newblock In {\em Proceedings of the IEEE/CVF Conference on Computer Vision and Pattern Recognition}, pages 14487--14496, 2024.

\bibitem{khurana2023point}
Tarasha Khurana, Peiyun Hu, David Held, and Deva Ramanan.
\newblock Point cloud forecasting as a proxy for 4d occupancy forecasting.
\newblock In {\em Proceedings of the IEEE/CVF Conference on Computer Vision and Pattern Recognition}, pages 1116--1124, 2023.

\bibitem{weng2022s2net}
Xinshuo Weng, Junyu Nan, Kuan-Hui Lee, Rowan McAllister, Adrien Gaidon, Nicholas Rhinehart, and Kris~M Kitani.
\newblock S2net: Stochastic sequential pointcloud forecasting.
\newblock In {\em European Conference on Computer Vision}, pages 549--564. Springer, 2022.

\bibitem{geiger2012we}
Andreas Geiger, Philip Lenz, and Raquel Urtasun.
\newblock Are we ready for autonomous driving? the kitti vision benchmark suite.
\newblock In {\em 2012 IEEE conference on computer vision and pattern recognition}, pages 3354--3361. IEEE, 2012.

\bibitem{caesar2020nuscenes}
Holger Caesar, Varun Bankiti, Alex~H Lang, Sourabh Vora, Venice~Erin Liong, Qiang Xu, Anush Krishnan, Yu~Pan, Giancarlo Baldan, and Oscar Beijbom.
\newblock nuscenes: A multimodal dataset for autonomous driving.
\newblock In {\em Proceedings of the IEEE/CVF conference on computer vision and pattern recognition}, pages 11621--11631, 2020.

\bibitem{peri2023towards}
Neehar Peri, Achal Dave, Deva Ramanan, and Shu Kong.
\newblock Towards long-tailed 3d detection.
\newblock In {\em Conference on Robot Learning}, pages 1904--1915. PMLR, 2023.

\bibitem{o2022predicting}
Amy~L O'Dell, Ashleigh~J Filtness, and Andrew~P Morris.
\newblock Predicting the intention of distracted pedestrians at road crossings.
\newblock {\em Accident Analysis \& Prevention}, 173:106707, 2022.

\bibitem{liu2024survey}
Mingyu Liu, Ekim Yurtsever, Jonathan Fossaert, Xingcheng Zhou, Walter Zimmer, Yuning Cui, Bare~Luka Zagar, and Alois~C Knoll.
\newblock A survey on autonomous driving datasets: Statistics, annotation quality, and a future outlook.
\newblock {\em IEEE Transactions on Intelligent Vehicles}, 2024.

\bibitem{amini2018variational}
Alexander Amini, Wilko Schwarting, Guy Rosman, Brandon Araki, Sertac Karaman, and Daniela Rus.
\newblock Variational autoencoder for end-to-end control of autonomous driving with novelty detection and training de-biasing.
\newblock In {\em 2018 IEEE/RSJ International Conference on Intelligent Robots and Systems (IROS)}, pages 568--575. IEEE, 2018.

\bibitem{greer2024towards}
Ross Greer and Mohan Trivedi.
\newblock Towards explainable, safe autonomous driving with language embeddings for novelty identification and active learning: Framework and experimental analysis with real-world data sets.
\newblock {\em arXiv preprint arXiv:2402.07320}, 2024.

\bibitem{Kulkarni2021}
Ninad Kulkarni, Akshay Rangesh, Jonathan Buck, Jeremy Feltracco, Mohan~M Trivedi, Nachiket Deo, Ross Greer, Saman Sarraf, and Suchitra Sathyanarayana.
\newblock Lisa amazonmlsl vehicle attributes (lava) dataset.
\newblock 2021.

\end{thebibliography}

\end{document}